%% file: paper.tex
\pdfoutput=1

\documentclass[11pt]{article}

\usepackage{acl}

\usepackage{times}
\usepackage{latexsym}

\usepackage{amssymb}

\usepackage[T1]{fontenc}

\usepackage[utf8]{inputenc}

\usepackage{microtype}

\usepackage{comment}

\usepackage{inconsolata}

\usepackage{graphicx}

\usepackage{soul}

\title{Introducing Bode: A Fine-Tuned Large Language Model for Portuguese Prompt-Based Task}

\author{Gabriel Lino Garcia$^1$, 
    Pedro Henrique Paiola$^{1}$, 
    Luis Henrique Morelli$^{1}$,
    Giovani Candido$^1$, \\
    \textbf{Arnaldo Cândido Júnior}$^2$,
    \textbf{Danilo Samuel Jodas}$^1$, 
    \textbf{Luis C. S. Afonso}$^1$,
    \textbf{Ivan Rizzo Guilherme}$^3$, \\ 
    \textbf{Bruno Elias Penteado}$^3$,
    \textbf{João Paulo Papa}$^1$ \\
    $^1$School of Sciences, São Paulo State University, Bauru, Brazil \\
    $^2$Institute of Natural Sciences and Technology, São Paulo State University, Rio Claro, Brazil \\
    $^3$Institute of Biosciences, Humanities and Exact Sciences, São Paulo State University, São José do Rio Preto, Brazil \\
    \small{\texttt{\{gabriel.lino, pedro.paiola, luis.morelli, giovani.candido, arnaldo.candido,}} \\
    \small{\texttt{ivan.guilherme, bruno.penteado, joao.papa\}@unesp.br, danilo.jodas@gmail.com}}
}

\begin{document}

\maketitle

\input{abstract}
\input{introduction}
\input{theoretical_background}
\input{related_works}
\input{proposed_model}

\input{methodology}
\input{experiments}

\input{conclusion}


\bibliography{references}




\end{document}

%% file: abstract.tex
\begin{abstract}
    Large Language Models (LLMs) are increasingly bringing advances to Natural Language Processing. However, low-resource languages, those lacking extensive prominence in datasets for various NLP tasks, or where existing datasets are not as substantial, such as Portuguese, already obtain several benefits from LLMs, but not to the same extent. LLMs trained on multilingual datasets normally struggle to respond to prompts in Portuguese satisfactorily, presenting, for example, code switching in their responses. This work proposes a fine-tuned LLaMA 2-based model for Portuguese prompts named Bode in two versions: 7B and 13B. We evaluate the performance of this model in classification tasks using the zero-shot approach with in-context learning, and compare it with other LLMs. Our main contribution is to bring an LLM with satisfactory results in the Portuguese language, as well as to provide a model that is free for research or commercial purposes.
\end{abstract}

%% file: introduction.tex
\section{Introduction}
\label{s.introduction}

Large Language Models (LLMs) have reshaped the landscape of how computers comprehend and generate text. These robust networks, trained on massive volumes of textual data, have become central partakers in Natural Language Processing (NLP) applications. Models such as GPT-$3$, developed by OpenAI \citep{brown2020language}, BERT, developed by Google \citep{devlin2019bert}, and LLaMA, developed by Meta \citep{touvron2023llama}, have demonstrated remarkable achievements in NLP tasks ranging from machine translation to text generation.

Language models have evolved to enhance the NLP capacities and cope with challenging aspects focused on high performance with fewer parameters and task-switching to particular applications based on better fine-tuning procedures. Recent architectures such as LLaMA $2$~\cite{touvron2023llama2}, Mistral $7$B~\cite{jiang2023mistral}, and Falcon $7$B~\cite{falcon40b} have supported increasing efficiency and accuracy while reducing the model's complexity and memory requirements for the training process. In addition, openly available LLMs have been highly demanded to drive research toward better generative architectures.

As the world has witnessed unprecedented progress in the capabilities of NLP models, particularly in English, the considerable computational costs associated with their development and training remain a significant challenge. Recent research, exemplified by the work of \citet{brown2020language} on GPT-$3$ and the comprehensive analysis conducted by \citet{devlin2019bert} on BERT, has shed insights on the massive scale of datasets and computing power required to design such models. This challenge has posed a significant obstacle to developing a more comprehensive range of high-quality models for less-spoken languages, such as Portuguese.

While multilingual models such as mBERT, proposed by \citet{devlin2019bert}, have expanded the scope of NLP applications, addressing the need for specific models for individual languages, such as Portuguese, is essential and required for particular applications. Despite their versatility, multilingual models may not capture the complex nuances and idiosyncrasies of a certain language.

Portuguese is spoken by over $260$ million people worldwide, making it a significant part of our global community. In this context, we adopt a feasible and efficient approach to develop a model for NLP tasks in Portuguese, known as Bode, a promising initiative to cope with this gap. The proposed model involves adapting and improving the prompt-based tasks by tailoring the LLaMA $2$ architecture for Portuguese instruction-following responses. By capitalizing on research over a publicly translated dataset of instruction-following tasks, the method applies a fine-tuning procedure to leverage the LLaMA $2$ capabilities to the context of the Portuguese language. Bode has the potential to stimulate research and innovation within the Portuguese-speaking community by providing customized solutions to address their unique singularities.

The structure of this paper can be summarized as follows: In Section$~\ref{s.theoretical_background}$, we offer a brief theoretical foundation for the proposed project. Following that, Section$~\ref{s.related_works}$ provides a concise review of related works, and Section$~\ref{s.proposed_model}$ introduces the proposed LLM. Then, Sections$~\ref{s.methodology}$ and$~\ref{s.experiments}$ outline the methodology and discuss the results yielded by the proposed model, respectively. To conclude, Section$~\ref{s.conclusion}$ presents our findings and overall conclusions of the work.

%% file: theoretical_background.tex
\section{Theoretical Background}
\label{s.theoretical_background}

This section presents the fundamental theoretical aspects that support the proposed research and is organized into subsections. Together, these subsections establish the essential aspects behind the methodology presented in this study.

\subsection{LLaMA Models}
\label{ss.llama}

LLaMA models \citep{touvron2023llama} are LLMs based on the well-established transformer architecture. These models borrow several training and inference mechanisms from other architectures, such as RMSNorm normalizing function~\citep{zhang2019root} for pre-normalization from GPT-$3$, the SwiGLU activation function~\citep{shazeer2020glu} from PaLM~\citep{chowdhery2022palm}, and the Rotary Position Embeddings~\citep{su2022roformer} from GPT-Neo~\citep{black2021gptneo} to enhance the training stability and improve the performance by reducing the number of hidden units. These models also employ causal multi-head attention to reduce memory usage and runtime by avoiding the storage of attention weights and the computation of key/query scores that are masked due to the causal nature of the language modeling task.

LLaMA $2$, introduced by \citet{touvron2023llama2}, is an updated iteration of the previous models. These new models were trained on a significantly expanded pre-training corpus, incorporating new publicly available data that extends it by over $40\%$. The primary architectural difference involves the introduction of Grouped-query Attention (GQA)~\citep{ainslie2023gqa}. GQA divides query heads into G groups, each sharing a single key and value head. This approach enables interpolation between Multi-head~\citep{vaswani2023attention} and Multi-query Attention~\citep{shazeer2019fast} while reducing the size of the key-value cache and the amount of data loaded, ultimately improving inference scalability. LLaMA 2 has three variants, with $7$, $13$, and $70$ billion parameters. Experiments have shown that LLaMA $2$ models outperform previous models on popular benchmarks.

\subsection{Mistral 7B}
\label{ss.mistral}

Mistral $7$B, proposed by \citet{jiang2023mistral}, is an LLM that focuses on high performance with efficient inference and reduced memory usage during decoding using GQA and Sliding Window Attention (SWA)~\citep{beltagy2020longformer}. Unlike traditional transformers, it optimizes performance by stacking layers and enabling access to information beyond the window size. A hidden state at position $i$ in layer $k$ can reach all the previous layer's hidden states within $i - W$, with $W$ representing the window size. Recursively, the hidden state can access tokens from the input layer up to a distance of $W \times k$. Thus, a fixed attention span allows limiting the cache size by utilizing a rolling buffer cache that prevents it from growing larger, and this reduction does not impact the model's quality negatively.

Lastly, predicting tokens individually during sequence generation is necessary, as each token depends on the preceding ones. However, the prompt is known in advance, enabling the pre-filling of the key-value cache by storing the previous prompt keys and values of transformers. This action streamlines the process, focusing solely on calculating the attention for the new token. In case of a very lengthy prompt, it is feasible to divide it into smaller segments and pre-fill the cache with each segment. The window size is selected as the chunk size, and attention is computed for each chunk, encompassing both the cache and the chunk itself.

Results have shown that Mistral $7$B outperforms LLaMA $2$ $13$B across various tasks. Additionally, it surpasses LLaMA $1 34$B on most benchmarks tested, demonstrating superior performance, particularly in code, mathematics, and reasoning.

\subsection{Falcon 7B}
\label{ss.falcon}

Falcon-$7$B, designed by Technology Innovation Institute (TII) \citep{falcon40b}, is an LLM pre-trained on $1.5$ trillion tokens from RefinedWeb \citep{penedo2023refinedweb}, which is composed of English web data. Its architecture, adapted from GPT-$3$, involves various modifications, comprising Rotary Position Embeddings, Multi-query Attention, and Flash Attention~\citep{dao2022flashattention}.

Multi-query Attention operates similarly to Multi-head Attention, allowing multiple heads to share a single set of keys and values. In addition, Flash Attention reduces the memory access during the precise attention calculation by performing Softmax reduction without considering the entire input and by avoiding storing intermediate attention matrices for the backward pass. Also, within the decoder block, there is the integration of parallel attention and a Multilayer Perceptron (MLP) featuring a single normalization layer. The model is available on Hugging Face\footnote{\url{https://huggingface.co/tiiuae/falcon-7b}}.


\subsection{Low-Rank Adaptation}
\label{ss.lora}

Low-Rank Adaptation (LoRA) is a technique designed to facilitate the adaptation of LLMs extensively pre-trained on a vast corpus of textual data to novel and unexplored tasks without incurring any inference latency or compromising the length of input sequences \citep{hu2021lora}. When confronted with a new downstream task, LoRA inserts a compact low-rank matrix into the pre-trained weight matrices of the language model. Then, these newly added parameters are trained with the specific downstream task and serve as the pivotal element for adapting the pre-trained weights to the unique requirements of the new task at hand.

\citet{hu2021lora} highlights the importance of tuning the rank of the low-rank matrix, which is a hyperparameter that can impact the number of trainable parameters and the model quality. In addition, by adapting the attention matrices and freezing the MLP modules for efficiency purposes, the authors demonstrated their strategy's potential by outperforming other adaptation approaches, such as the default fine-tuning process.

Compared to the standard fine-tuning procedure, LoRA significantly reduces the trainable parameters required for adaptation to downstream tasks while maintaining a high-quality model. Another valuable feature of LoRA concerns its ability to facilitate quick task-switching when deployed as a service, as it allows for sharing most model parameters.

%% file: related_works.tex
\section{Related Works}
\label{s.related_works}

In this section, we review two notable contributions in the context of NLP applications, with a particular emphasis on the Portuguese language. Both works present different strategies for performing monolingual pre-training in Portuguese to enhance models extensively trained on English-centric text corpora. The subsection $\ref{ss.sabia}$ discusses the work of \citet{pires2023sabia}, which introduces Sabiá $7$B. The subsection $\ref{ss.cabrita}$ explores the research conducted by \citet{larcher2023cabrita}, which proposes openCabrita $3$B.

\subsection{Sabiá}
\label{ss.sabia}

Sabiá, introduced by Maritaca AI \citep{pires2023sabia}, challenges the practice of pre-training a single model for multiple languages. Their research shows that monolingual pre-training in the target language, Portuguese in this case, significantly improves models that have undergone extensive training on a diverse text corpus in English. As a result, they introduced Sabiá, a Portuguese LLM derived from further pre-training GPT-J~\citep{wang2021gptj} and LLaMA models on Portuguese-specific texts. Sabiá models were pre-trained using Portuguese subsets from the ClueWeb $2022$~\citep{overwijk2022clueweb} dataset with LLaMA's $7$ billion and $65$ billion parameters architectures.

Conversely, Sabiá-J models used GPT-J with $6$ billion parameters, following GPT-$3$'s architecture and hyperparameters. However, these models underwent 18 days of training on $7.8$ billion tokens, roughly one epoch of the Portuguese dataset, resulting in fewer tokens trained than Sabiá.

Both models were assessed on the Portuguese Evaluation Tasks (Poeta) benchmark, which includes $14$ NLP datasets in Portuguese. This evaluation followed a few-shot approach, where specific examples were manually selected for each dataset, limiting them to fit within a $2048$-token context. The results showcased the advantages of language-specific specialization using a nearly state-of-the-art $65$-billion-parameter model for Portuguese.

It is essential to note that Sabiá is not openly available to the public. Access to the model may require arrangements with Maritaca AI.

\subsection{Cabrita}
\label{ss.cabrita}

In a recent work, \citet{larcher2023cabrita} introduced openCabrita $3$B, an LLM designed for NLP tasks involving the Portuguese language. This model relies on the $3$-billion-parameter version of OpenLLaMA~\citep{openlm2023openllama}, which reproduces the widely known Meta's LLaMA $3$B, a state-of-the-art LLM pre-trained on $1$ trillion English tokens.

To enhance its ability to handle the Portuguese language effectively, openCabrita $3$B underwent a process known as Tokenizer Adaptation involving two steps. At first, a new tokenizer was trained on a corpus of Portuguese texts sourced from Wikipedia. Subsequently, the newly crafted tokenizer was integrated with OpenLLaMA's default tokenizer. Notably, new tokens were appended to the end of the tokenizer matrix, resulting in a combined tokenizer matrix comprising $52,000$ tokens. The outcome yielded a bilingual tokenizer capable of handling English and Portuguese.

After the adaptation strategy, the original script and hyperparameters from the OpenLLaMA training were employed for both pre-training and training of openCabrita $3$B. In the preprocessing phase, a Portuguese subset of the mC$4$ dataset~\citep{xue2021mt5} was filtered to retain only the highest-quality documents. From this filtered dataset, $7$ billion tokens were used in the continued pre-training stage. For training purposes, $170$ billion tokens were employed.

The openCabrita $3$B demonstrated commendable performance on a range of Portuguese NLP tasks, including Question-Answering (QA), classification, and regression, consistently outperforming openLLaMA $3$B. It closely rivaled the openCabrita version that relies only on the default tokenizer without adaptation, occasionally surpassing it in some scenarios. Moreover, openCabrita $3$B closely approaches the performance of Sabiá-J ($6$B), demonstrating superiority over GPT-J ($6$B) and competitive performance compared to LLaMA $7$B.


%% file: proposed_model.tex
\section{Proposed Model}
\label{s.proposed_model}

\begin{figure*}[!ht]
    \centering
    \includegraphics[width=0.7\textwidth]{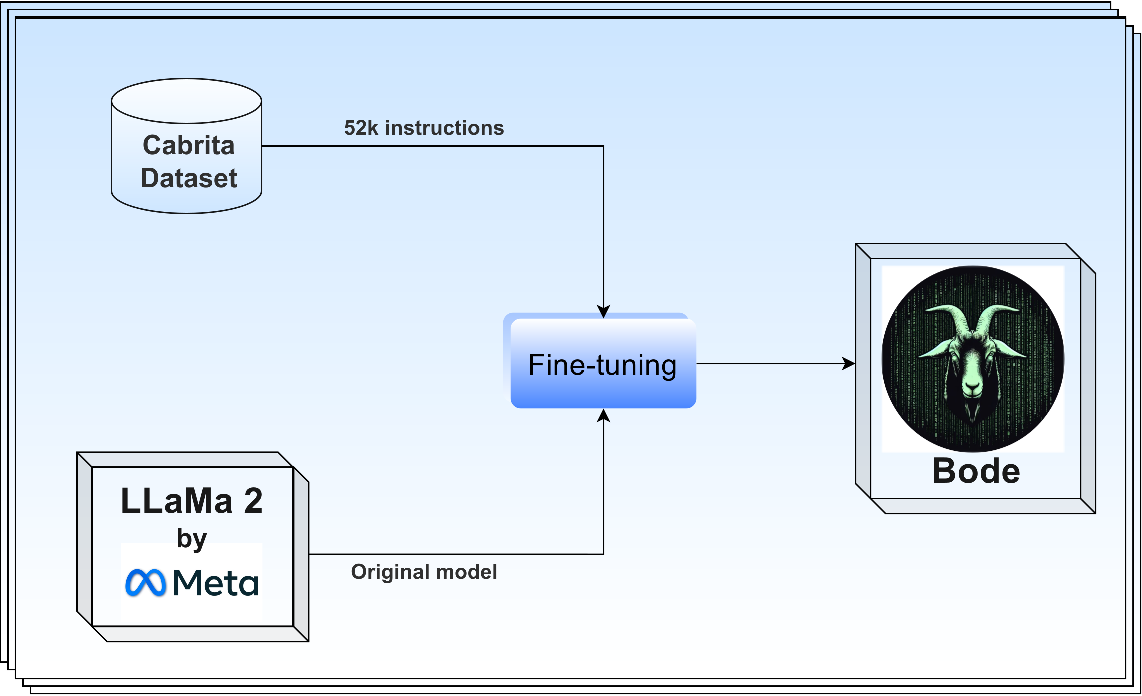}
    \caption[Overview of the proposed method.]{Overview of the proposed method.\footnotemark}
    \label{f.bode_model}
\end{figure*}

The proposed method aims to provide a fine-tuned language model tailored for Portuguese prompt response applications based on the new LLaMA $2$ architecture. Figure~\ref{f.bode_model} depicts the overview of the fine-tuning approach employed in this work. Similar to Cabrita, the proposed method is designed to harness the capability of the LLaMA model to address instruction-following tasks by using a large dataset with a broad range of Portuguese instructions. However, we extend the counterpart work by leveraging the newly designed version of the LLM called LLaMA $2$ by Meta AI. This new architecture provides several improvements ranging from a larger number of parameters to a more human-machine interaction supported by training on additional data through reinforcement learning. This process may lead to better interactive, reliable communication and more accurate response generation.

In terms of the fine-tuning procedure, the base LLaMA $2$ model was trained on the same Portuguese dataset assembled by the work of \citet{larcher2023cabrita}, which provides a broad range of Portuguese tokens. The fine-tuning procedure to create the Bode model was performed using the Low-Rank Adaptation method based on the following hyperparameter values: i) LoRA alpha is set to $32$, and ii) the dropout rate is set to $0.05$.

\footnotetext{The image was created by the authors, and the Bode logo was generated using Bing Chat. ``Meta-Logo'' by doravand is licensed under CC BY 2.0. To view a copy of this license, visit \url{https://creativecommons.org/licenses/by/2.0/?ref=openverse.}.}

%% file: methodology.tex
\section{Methodology}
\label{s.methodology}

This section describes the methodology employed to support the evaluation of the proposed model in terms of its capability to generate accurate Portuguese responses compared to other LLMs. We provide an overview of the experimental setup, the description of the zero-shot and in-context-learning approaches, and datasets used for evaluation.

\subsection{Experimental Setup}

The experiments for news classification tasks were conducted on the supercomputer Santos Dumont (SDumont)\footnote{\url{https://sdumont.lncc.br/machine.php}}, which boasts a processing capacity of $5.1$ PetaFLOPS. It features a hybrid configuration of computational nodes, leveraging available parallel processing architecture. The node used was the Bull Sequana X$1120$, equipped with $2$ Intel Xeon Skylake $6252$ CPUs with $24$ cores each, totaling $48$ cores, $384$GB of RAM, and $4$ NVIDIA Volta V$100$ GPUs. Additionally, the Kaggle platform was utilized for the sentiment analysis tasks, which provided access to an environment equipped with an NVIDIA K$80$ GPU. The code was written in Python $3.10$ with PyTorch $2.0.0$ and Hugging Face. We adopted zero-shot and in-context learning approaches for the LLM's learning. Each approach is further explained in the following sections.

\subsubsection{Zero-Shot Learning} 
In zero-shot learning, the proposed model is requested to produce an output for previously untrained data. This process involves using a transfer learning approach by harnessing a pre-trained model for which the input data labels were unavailable during the training time~\cite{tan2021survey}. This procedure instructs the model to generalize to unexplored scenarios. For the zero-shot approach, experiments were conducted without performing any fine-tuning procedure in the Bode model for the given classification task.

\subsubsection{In-Context Learning} 
The in-context learning approach is designed to explore the model's capability regarding tasks related to specific contextual information. According to~\cite{brown2020language}, contextual learning is employed to adapt the models to the context provided by the input data, which is assembled from a similar task for which the model has been trained previously, thus providing the ability to produce more accurate responses while refraining from a further training of the model. Unlike the classical fine-tuning procedure, which enables the models to change the learning to a different task or domain by updating their weights, in-context learning allows the harnessing of the same context provided by massive amounts of data to promote slight or moderate adjustments to the same contextual domain the models have been introduced previously. Therefore, we employed contextual learning in the Bode model to promote improvements and provide a better performance according to the context of the dataset used.

\subsubsection{Prompt Engineering}
Prompt Engineering is a relatively recent method designed to promote adjustment to LLM models by providing new contexts and specific instructions during human-machine communication. This process involves articulating new prompts for specific contexts and using them to assist the model in comprehending the task or domain to promote adaptation in generating more detailed responses. This method is essential for either zero-shot or in-context learning to support adjustments to different tasks.

In the context of this work, experiments conducted on prompt engineering applications were designed to expose the proposed model to a broad range of classification tasks and support fine adjustments intended to maximize its accuracy. For a broader assessment, we provide two prompt applications exhibiting the model's capacity to respond to sentiment analysis tasks and news category classification in terms of the multiclass classification analysis. The following samples illustrate the prompt applications in the two classification tasks:

\begin{itemize}
    \item Sentiment analysis into 3 classes: 
    \subitem Prompt: \textit{``Você é um assistente de perguntas e respostas. Cada contexto passado será um tweet que está vinculado a um sentimento correspondente. No total, são 3 tipos de sentimentos: Positivo, Neutro e Negativo. O seu objetivo é dado um tweet, encontrar qual é o seu sentimento correspondente. Abaixo estão alguns exemplos:}
    
    \textit{Tweet: :D que lindo dia ! Resposta: Positivo}
    
    \textit{Tweet: eu tô tão cansado :( Resposta: Negativo}
    
    \textit{Tweet: Microsoft lança pesquisa resultado de pesquisa com o Ibope sobre uso da \#tecnologia no \#trabalho no \#Brasil @MicrosoftBr \@NielsenIBOPE Resposta: Neutro}
    
    \textit{Considere que tweets com chamadas de notícias com sempre neutros, independente do seu conteúdo, a não ser que o autor emita sua opinião sobre o acontecimento relatado. Dado o contexto, responda em qual dos 3 tipos de sentimentos o tweet a seguir se enquadra.''}

   \item News category classification into 4 classes:
    \subitem Prompt: \textit{"Você é um assistente de perguntas e respostas. Cada contexto passado será uma notícia que está vinculada a uma categoria correspondente. No total, são 4 categorias: Mundo, Esportes, Negócios e Tecnologia. O seu objetivo é dado uma notícia, encontrar qual é a sua categoria correspondente. Abaixo estão alguns exemplos:}
    
    \textit{Notícia: Até o final do ano, a gigante da computação planeja ter seu maior número de funcionários desde 1991. Resposta: Tecnologia}
    
   \textit{ Notícia: Michael Phelps ganhou a medalha de ouro no 400 Medley individual e estabeleceu um recorde mundial em um tempo de 4 minutos 8,26 segundos. Resposta: Esportes}
    
    \textit{Notícia: TEEHRAN (Reuters) - Um oficial militar iraniano disse domingo a Israel e os Estados Unidos não ousariam atacar o Irã, pois poderia revidar em qualquer lugar de Israel com seus mais recentes mísseis, informou as agências de notícias. Resposta: Mundo}
    
    \textit{Notícia: Reuters-Os varejistas de vestuário esperam que suas modas de volta às aulas façam a nota entre os adolescentes conscientes do estilo e os jovens adultos neste outono, mas pode ser uma venda difícil, com os alunos e os pais mantendo um mais apertado. carteiras. Resposta: Negócios}
    
    \textit{Dado o contexto, responda em qual das 4 categorias a notícia a seguir se enquadra.}"
\end{itemize}

For a given prompt, the model's output encompasses the content category and additional text comprising a brief description of the input prompt. Therefore, after extensive empirical observations, we realized that the first word of the generated content is usually considered the correct class for the given text.

\subsection{Baseline models}

The performance evaluation was assessed by comparing the proposed Bode model with the following existing state-of-the-art large language architectures:

\begin{enumerate}
    \item Cabrita $7$B, and openCabrita $3$B~\cite{larcher2023cabrita};
    \item Falcon $7$B~\cite{falcon40b};
    \item Llama $2$ $7$B, and Llama $2$ $13$B~\cite{touvron2023llama2};
    \item Mistral $7$B~\cite{jiang2023mistral}.
\end{enumerate}

The primary evaluation is concerned with comparing Cabrita $7$B with the proposed Bode model based on the LaMMA $2$ architecture. The authors designed a Cabrita $7$B\footnote{\url{https://huggingface.co/22h/cabrita-lora-v0-1}} model based on the first version of the LaMMA architecture. However, at the time of the writing of this work, there was no paper about it, and the underlying LLaMA model was unavailable\footnote{\url{https://huggingface.co/decapoda-research/llama-7b-hf}}. Therefore, due to such unexpected constraints, we restricted the assessment to the openCabrita $3$B model proposed in the work of \citet{larcher2023cabrita}.

\subsection{Datasets for evaluation}
The experiments were performed over three Portuguese datasets to assist in evaluating the proposed model in different real-world applications. Table~\ref{t.fact} exhibits the used datasets in detail, including their related classification task and the number of samples adopted in each experiment. Moreover, the subjective aspect of instruction-following datasets in terms of performance evaluation and quality of the given responses is worth noting. In this sense, the research focus hinges on classification tasks.

\begin{table}[!hbt]
    \centering
	\caption{Datasets employed in each experiment.}\label{t.fact}
		\begin{tabular}{ccc}
			\hline
			\textbf{Dataset}                 & \textbf{Classification Task} & \textbf{Samples}\\
			\hline
			\textit{TweetSentBr}               & Multiclass & 4,999\\
			\textit{TweetSentBr}               & Binary & 5,000  \\
			\textit{AGNews}                    & Multiclass & 7,600\\
			\textit{FakeRecogna}               & Binary & 10,560\\
			\hline
	\end{tabular}
\end{table}

The following items provide a detailed description of each dataset:
\begin{itemize}
	\item \textit{TweetSentBr}\footnote{\url{https://www.kaggle.com/datasets/augustop/portuguese-tweets-for-sentiment-analysis}}: A Portuguese sentiment analysis dataset assembled from Twitter content~\cite{brum2017building}. The data can be tasked with multiclass or binary classification problems, with the multiclass data labeled into positive, negative, and neutral classes. We opted to use only the test set supplied with the dataset to avoid the computational expenses associated with the model's training. Therefore, only $5,000$ samples were used in this phase.
	\item \textit{AGNews}\footnote{\url{https://www.kaggle.com/datasets/amananandrai/ag-news-classification-dataset}}: A dataset comprising a collection of news articles evenly distributed among four news categories, thus promoting a multiclass classification task. Due to the computational workload, we adopted only the test set for the model's training phase, thus leading to $7,600$ samples.
	\item \textit{FakeRecogna}\footnote{\url{https://github.com/Gabriel-Lino-Garcia/FakeRecogna}}: A dataset designed for fake news detection purposes, with samples evenly distributed across real and fake news categories~\cite{garcia2022fakerecogna}. The version used in this study was assembled from the test set of the original FakeRecogna dataset, thus leading to $5,280$ samples in each class of news articles.
\end{itemize}

All experiments were conducted in the test samples provided by their respective datasets. However, those samples are also representative since they originate from the same data distribution of the entire dataset. By leveraging the provided test set, we reduced the computational workload required during the model's evaluation, resulting in a more efficient and cost-effective approach rather than using the entire test dataset in the model's performance assessment.

\subsection{Evaluation Metrics}

We adopted the evaluation procedure by using only the test sets from each dataset. It is worth noting that the fine-tuning procedure was only performed using the openCabrita Portuguese dataset to train the proposed Bode model with the LaMMA $2$ architecture. Subsequently, the performance of the models was evaluated by calculating the average values for accuracy and F1-score over the test sets of each dataset using zero-shot and in-context learning without any additional training with the baseline Portuguese datasets.

%% file: experiments.tex
\section{Experiments and results}
\label{s.experiments}

Before presenting the results and discussing the experiments conducted in this article, it is important to emphasize that we did not include the results of Cabrita $7$B in all experiments due to its unavailability during the preparation of this work. However, we chose to retain the results that had already been obtained, as Cabrita $7$B was one of the few LLMs trained and freely available for the Portuguese language. Furthermore, there is the prospect that the authors may address the availability issue of the model in the near future. 

\begin{figure*}[!ht]
    \centering
    \includegraphics[width=1.0\textwidth]{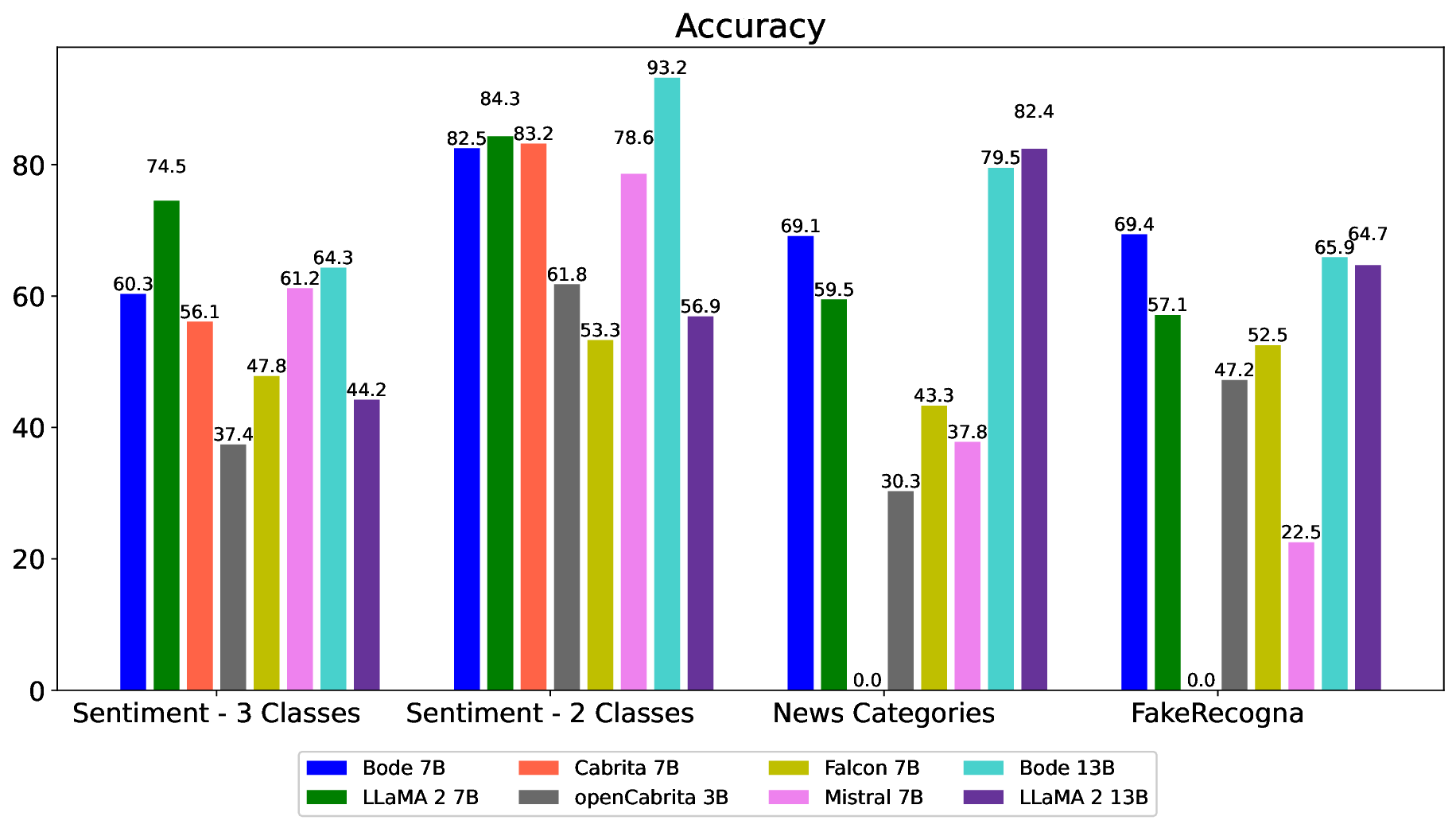}
    \caption{Accuracy values for each dataset and LLM adopted in the experiments.}
    \label{f.accuracy}
\end{figure*}

\begin{figure*}[!ht]
    \centering
    \includegraphics[width=1.0\textwidth]{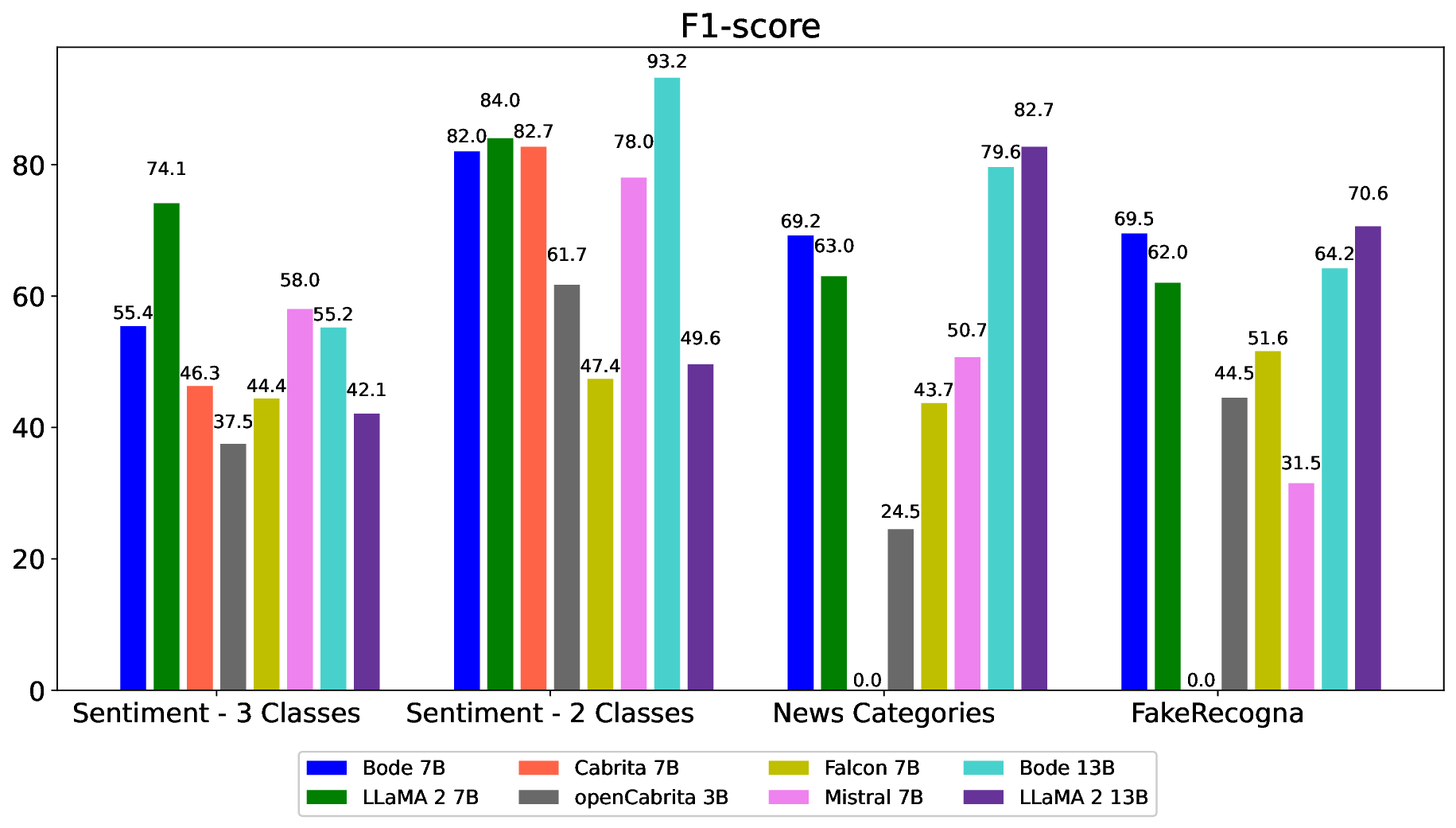}
    \caption{F1-score values for each dataset and LLM adopted in the experiments.}
    \label{f.f1}
\end{figure*}

As previously alluded to, a total of four experiments were conducted across a range of tasks and datasets. \autoref{f.accuracy} illustrates the outcomes of the models across the various tasks, employing accuracy as the primary evaluative metric. Furthermore, in accordance with the depiction provided in \autoref{f.f1}, emphasis on the results obtained under the evaluation metric of F1-score.


In this context, it becomes evident that, for sentiment analysis tasks, whether with two or three classes, the LLaMa $2$ $7$B showed superior performance compared to the other base models with the same number of parameters. In contrast, Bode $13$B demonstrated higher evaluation scores than the base models with $13$B parameters, with a substantial advantage over other LLMs in this case. It is worth noting that Bode $13$B achieved an accuracy of over $90$\%, while its counterpart version with the same number of parameters, LLaMa 2 $13$B, achieved only $56.9$\%. When considering multi-class news classification, Bode $7$B exhibits the highest performance among the $7$B models, whereas a marginal difference of $0.9$\% is observed in favor of LLaMa 2 $13$B among the $13$B models. Finally, in the context of fake news identification, both Bode models outperform their counterparts, underscoring the consistency of the Bode model's performance, which consistently ranks among the top performers, even though it may not excel in every single task.

Considering the Bode model and its corresponding base model, LLaMa $2$ $7$B, it becomes evident that both exhibit optimal configurations that vary depending on the specific task at hand. However, one plausible explanation for the observed performance decrement of the Bode model in comparison to LLaMa $2$ $7$B may be attributed to catastrophic forgetting. This phenomenon pertains to the possibility that the model may inadvertently unlearn previously acquired knowledge during fine-tuning, consequently leading to the loss of critical information.







%% file: conclusion.tex
\section{Conclusion}
\label{s.conclusion}

In this article, we introduce Bode, a large-scale language model developed to advance the field of NLP in the Portuguese language. Bode is built upon the LLaMA $2$ model, incorporating elements from the $7$B and $13$B models, adapted for the context of the Portuguese language. Language models like Bode play a crucial role in a wide range of applications, from virtual assistants to machine translation systems, and have a significant impact on academic research.

During the training phase of Bode, we employed the Portuguese-translated version of Alpaca as our training dataset. This process was instrumental in adapting the model to the language and specific grammar of Portuguese, enabling Bode to comprehend and generate text within a more culturally and linguistically appropriate context. 

The results achieved by Bode in binary and multi-class classification tasks were promising and demonstrated the model's effectiveness. This outcome suggests that Bode has the potential to enhance a variety of NLP applications in Portuguese, from sentiment analysis to news categorization.

We hope this work will encourage other researchers and developers to explore further and enhance the model, thereby contributing to the continuous advancement of NLP in the Portuguese language. However, the ongoing development of Bode is essential, and we aim to enhance it over time by expanding its training data and refining its performance across a range of tasks. 